\pdfoutput=1

\documentclass[11pt]{article}

\usepackage[]{ACL2023}

\usepackage{times}
\usepackage{latexsym}

\usepackage[T1]{fontenc}

\usepackage[utf8]{inputenc}

\usepackage{microtype}

\usepackage{graphicx}
\usepackage{inconsolata}
\usepackage{algorithm}
\usepackage{algorithmic}
\usepackage{amsmath}
\usepackage{amssymb}
\usepackage{color}
\usepackage{multirow}
\usepackage{array}
\newcolumntype{M}[1]{>{\centering\arraybackslash}m{#1}}
\usepackage{booktabs}

\title{Balancing Effect of Training Dataset Distribution of Multiple Styles for Multi-Style Text Transfer}

\author{Debarati Das \quad David Ma \quad Dongyeop Kang \\
         Department of Computer Science, University of Minnesota \\
                 \texttt{\{das00015, maxxx818, dongyeop}\}@umn.edu\\
         }

\begin{document}
\maketitle
\begin{abstract}
Text style transfer is an exciting task within the field of natural language generation that is often plagued by the need for high-quality paired datasets. Furthermore, training a model for multi-attribute text style transfer requires datasets with sufficient support across all combinations of the considered stylistic attributes, adding to the challenges of training a style transfer model. 
This paper explores the impact of training data input diversity on the quality of the generated text from the multi-style transfer model. We construct a pseudo-parallel dataset by devising heuristics to adjust the style distribution in the training samples. We balance our training dataset using marginal and joint distributions to train our style transfer models. We observe that a balanced dataset produces more effective control effects over multiple styles than an imbalanced or skewed one. Through quantitative analysis, we explore the impact of multiple style distributions in training data on style-transferred output. These findings will better inform the design of style-transfer datasets.
\end{abstract}

\section{Introduction}

Multi-style text transfer is a challenging task today with applications such as automatic domain-appropriate, style-conformant writing \cite{fu2018style} and AI-assisted stylistic language editing. Text style transfer is an intricate task as all language has a specific context, and those contexts influence the attributes of the language \cite{hovy2021importance}. Text style transfer is challenging because it involves dealing with the aspects of style coupled with the textual content \cite{hu2017toward, shen2017style, lample2018multiple}. This domain's other obstacles include the need for parallel corpus \cite{jhamtani2017shakespearizing} and quality training data. As the number of style dimensions increases with multi-style text transfer, not only is the requirement of a jointly annotated corpus across all the stylistic dimensions problematic, but the different styles are not necessarily independent.\\
\begin{figure}[t]
\centering
\includegraphics[width=0.52\textwidth]{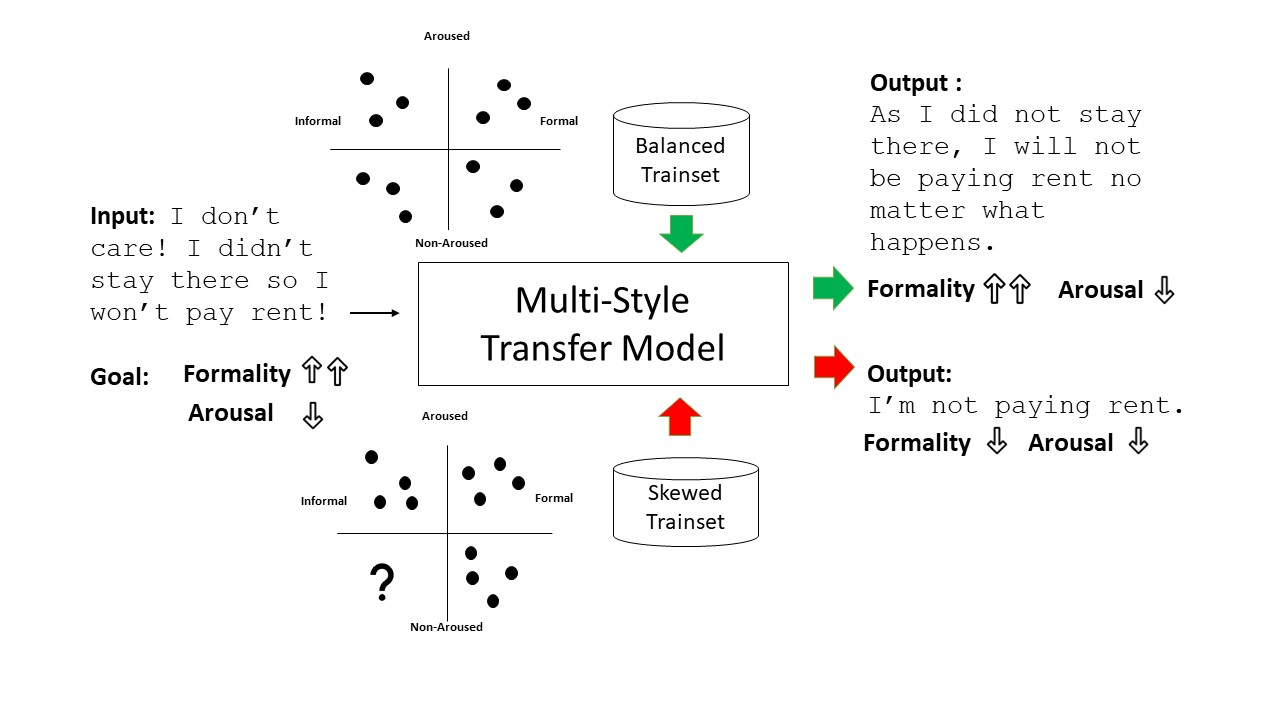} 
\caption{When an input sentence is passed to the multi-style transfer model, to increase formality and decrease arousal, we hypothesize that when the model is trained on a \textit{balanced} joint distribution of formality and arousal (all four style combinations have a 25\% representation) - the style transfer is more successful as opposed to when the model is trained on a \textit{skewed} joint distribution (there is no representation of the ``informal unaroused'' style combination) of styles in the training data.}
\label{fig:fig1}
\end{figure}
\begin{figure*}[t]
\centering
\includegraphics[width=0.8\textwidth,trim={0.3cm 2.3cm 1cm 0}, clip]{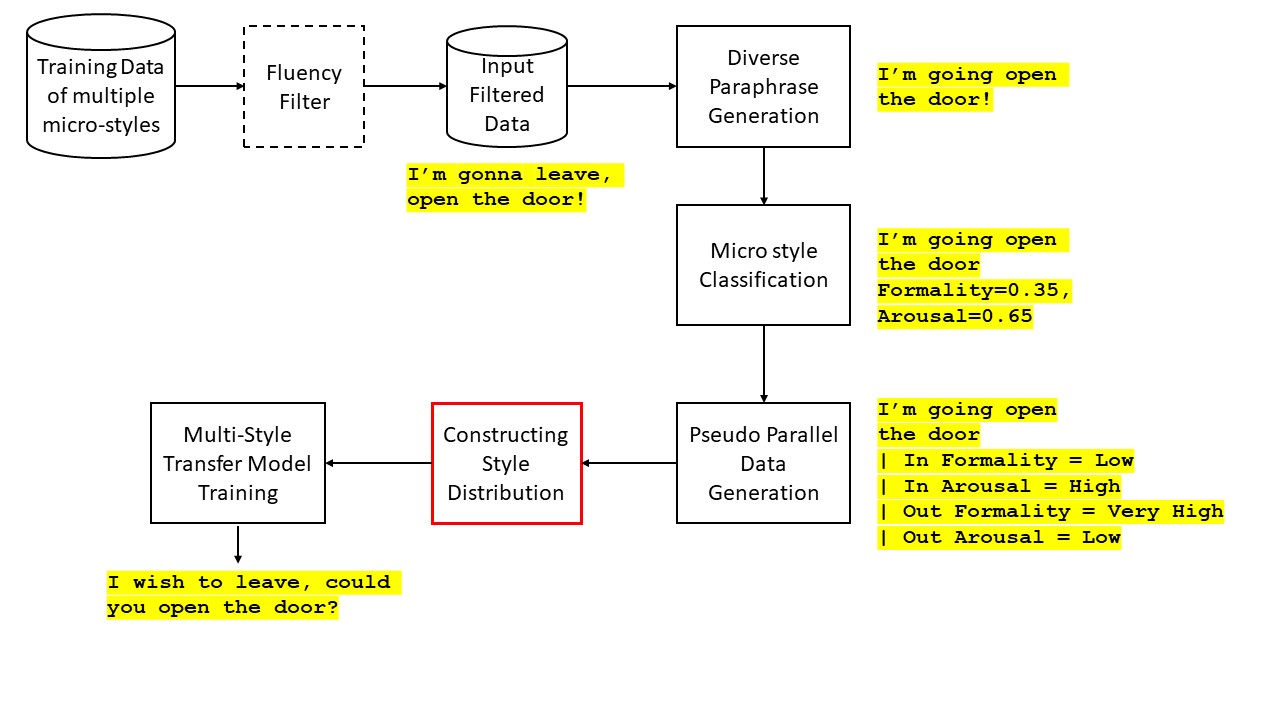} 
\caption{The input sentence transitions through every step in our multi-style text style transfer pipeline. The box in red indicates our main contribution to the pipeline, which helps us explore the effects of joint micro-style combinations on style-transferred output.}
\label{fig:fig2}
\end{figure*}
While ``style'' can also refer to authorial or domain-specific style, in this paper, we focus on ``micro-styles'' as defined by \cite{kang2021style} where they define ``micro-style'' as a complex combination of different factors such as formality markers, emotions, and metaphors. People intentionally \cite{troiano2021theories} tune these styles in writing differently based on their mood, the person they are addressing, the content of the message, or the platform. Multiple micro-styles can jointly describe a text; for example, a given text could simultaneously be formal and sad. Micro-styles also more easily lend themselves to being represented as spectra with varying degrees of intensity. These points align with our vision of an application where users can edit micro-style aspects of their writing.

Much research exists on models implementing multi-style text transfer and interdependency of micro-styles \cite{kang2019xslue, goyal2020multi, subramanian2018multiple}. However, there needs to be more exploration of the joint distribution of inherent micro-styles in the style transfer training dataset and how these micro-style distributions are related. Therefore, we pose a question - \textit{Can a dataset with minimal variance across multiple micro-style combinations, such that it experiences a ``balancing effect'', lead to a better style transferred output ?} Figure~\ref{fig:fig1} illustrates our intuition that a dataset that experiences a ``balancing effect'' will have more control over the multi-style transferred output than a ``skewed'' dataset. Suppose the style transfer model sees examples of every style combination that can exist - this could aid in the style generation of even unlikely combinations of styles compared to a skewed distribution of these joint micro-styles.

In this research, we consider a multi-style text style transfer pipeline assuming that the user has no access to parallel data or the style of the original text that he wishes to transfer, as would seem natural for a style language editing application. We introduce the changing of the training dataset micro-style joint distributions in such a pipeline and quantitatively explore the impact of this modification on the style transferred output. We perform a set of empirical analyses to demonstrate the influence of joint distributions on style-transferred output and show how this trend varies as the number of micro-styles considered changes. The `balancing effect' on a training dataset leads to style transferred sentences from even the joint style combinations that are typically rare (``informal unbiased and unaroused''). Our study is the first of its kind on the distribution of micro styles in training datasets for multi-style text style transfer and is likely to have implications for designing datasets for multi-style transfer model training and fall within the context of and align with recent work on characterizing datasets and factors impacting style transfer \cite{bender2018data,schoch2021contextualizing, li2019domain, zhang2020parallel, gururangan2018annotation}. 

\section {Multi Style Transfer Pipeline}

\textbf{Datasets:} We chose four micro-styles from the style hierarchy defined in \citeauthor{troiano2021theories}: Formality, Arousal, Sentiment, and Bias, for our study and used publicly available NLP datasets built by other researchers \cite{rao2018dear, buechel2022emobank, go2009twitter, pryzant2020automatically,kang2019xslue} to develop and test our models. Appendix \ref{sec:dataset} mentions the details of the datasets and their usage.\\
\noindent \textbf{Pipeline Overview:} Our experimental setup for multi-style transfer is inspired by the work of \cite{krishna2020reformulating}. Like them, we first generate a ``diverse'' paraphrase of the input sentence, and then the paraphrased sentence is rewritten in the style of choice. Towards this end, we train a paraphrasing model (separately on a parallel paraphrase dataset). Then, the trained paraphrase model is used to create ``pseudo-gold'' parallel data for training style models. 

First, we adopted a pre-trained T5 model \cite{raffel2020exploring} to generate paraphrases. This model was trained for the task of paraphrase generation on the ParaNMT-filtered dataset provided by \cite{krishna2020reformulating}. Once we had this trained paraphrase model, we used diverse beam search \cite{vijayakumar2016diverse} to generate diverse fluent paraphrased outputs. An important assumption is that the paraphrase is stripped of its original style and does not leak into the training.  

We address this potential issue by training classifiers \cite{sanh2019distilbert} to predict style on the original and paraphrased datasets and find that all our micro-style classifiers have a classification accuracy of higher than 80\% F1, which is acceptable for pseudo-label creation. After we generate diverse paraphrases, we choose the most diverse paraphrase and then derive micro-style classifications for the paraphrased sentence using our trained micro-style classifiers. Therefore each sentence is assigned a classification score for each micro-style label and can form a "pseudo parallel” dataset for training the T5-based joint transfer model. Thus, our approach does not need a parallel dataset. 

We then converted the classifier predictions into buckets of style (ranging from ``very low'' to ``very high'') based on the chosen style of the original and then paraphrased sentences. The bucketing process is described in Appendix \ref{sec:app1}. After this step, we introduce our contribution of ``constructing style distributions" into the pipeline, as illustrated in Figure~\ref{fig:fig2}. Following that, we perform multi-style text style transfer. We appended the ``bucket'' information to the paraphrased sentence to achieve the necessary intensity transfers, as motivated by the original T5 paper \cite{raffel2020exploring}. We train T5-based style transfer models, where the paraphrased sentence and its style buckets are used as input parameters, while the style buckets assigned to the anchor sentence are used as proxy levels of output style transfer. All model-specific details are provided in Appendix \ref{sec:app1}. For generating sentences from our trained models, we used beam search \cite{vijayakumar2016diverse} and nucleus sampling \cite{holtzman2019curious} and chose the top 3 sentences from the generations. The following is an example of the input to the joint style transfer model and the expected output. \\

\noindent
\texttt{Goal} - Highly increase the formality of the sentence, slightly increase the arousal of the sentence \\
\texttt{Input} - transfer: I’m sad you’re going $\vert$ input formality: low $\vert$ input arousal: low $\vert$ output formality: high $\vert$ output arousal: mid \\
\texttt{Output} - I am sorry you are going to go.\\

Thus, we implemented a multi-style transfer pipeline to test our hypothesis without any finicky modeling paradigms popular in style transfer research, such as variational inference or autoregressive sampling \cite{he2020probabilistic, subramanian2018multiple}. 
\begin{table}[H]
\centering

\begin{tabular}{M{3.8cm} M{1.2cm} M{1cm}}
\toprule
 \textbf{Style Combination} & \textbf{Balanced} & \textbf{Skewed} \\
 \hline
  
  Formal Aroused & 3395 & 8685\\
  Formal Unaroused & 3395 & 2792 \\
  Informal Aroused & 3395 & 1275\\
  Informal Unaroused & 3395 & 828\\
  \bottomrule
\end{tabular}

 \caption{\label{tab:table0} Training data statistics (number of samples) for the balanced and skewed settings, when considering the micro-styles of Formality and Arousal.}
\end{table}

\noindent \textbf{Constructing Micro-style Distributions}
We define a ``style combination'' as a possible combination of the states that the micro-styles can take together - such as 'informal biased negative.' Since there are three micro-styles, each having binary states, the total possible number of style combinations, in this case, is given by $N_c = 2 \times 2 \times 2 = 2^3$. 
Therefore to generalize, if $\vert m_i \vert$ indicates the cardinality of each micro-style and $n$ indicates the number of micro-styles considered, the total possible number of style combinations ($N_c$) possible is given by : 
\begin{equation}
N_c = \prod_{i=1}^{n}  \vert m_i \vert
\end{equation}

To create the \textbf{balanced} joint distribution of styles, we ensure the standard deviation across the style combinations is close to 0. We do this by down-sampling each style combination, such that the number of samples in each style combination is the same as the least represented style combination. As we increase micro-styles, some micro-style combinations do not occur naturally together, so their representation is close to 0. In such cases, we assume that the least represented style combination is at least 5\% of the total dataset. To ensure our comparison across the ``balanced'' and ``skew'' settings is fair, we construct a \textbf{skewed} dataset with a total sample size that is the same as that of the balanced dataset. Thus, the balanced dataset has a uniform distribution, while the skewed dataset has a non-uniform distribution. Table \ref{tab:table0} shows the number of samples in each style combination of Formality and Arousal, given a ``balanced`` and ``skewed`` setting.

\begin{table*}[ht]
\scriptsize
\begin{tabular}{p{0.14\linewidth}p{0.29\linewidth}p{0.26\linewidth}p{0.21\linewidth}}
\hline
\textbf{Style Transfer Goal} & \textbf{Input Text} & \textbf{Balanced Transferred Text} & \textbf{Skewed Transferred Text}\\
\hline
$\uparrow$ Formality $\uparrow$ Arousal & Wouldn't it be great if Trump went 3rd party and sucked away millions of Republican votes lol & It wouldn't be nice if Trump went to the third party and swooped millions of Republican votes. & Would it not be great if Trump went 3rd party and sucked away millions of Republican votes? \\
\hline
$\uparrow$ Formality $\downarrow$ Arousal & I didn't know what happiness was until I got married. But by then it was too late. & Until I got married, I didn't even know what happiness was. & I did not know what happiness was till I got married and it was too late.\\
\hline
$\downarrow$ Formality $\uparrow$ Arousal & Did you hear about the soldier with 8 limbs? He was army & He's an army soldier with 8 legs? & Did you hear about the soldier with 8 limbs in the army? \\
\hline
$\downarrow$ Formality $\downarrow$ Arousal & Yeah, I don't understand all the hate. & Yeah I'm not gonna understand the hate. & Yeah I do not understand all the hate. \\
\hline
\end{tabular}
\caption{The table shows the style transferred sentences, given an input sentence and the intended style transfer goal, for both the balanced setting as well as the skewed setting.}
\label{tab:qual}
\end{table*}

\section{Experimental Results and Discussion}

\textbf{Evaluation Metrics:} Style transfer accuracy metrics quantify how nicely output texts match the desired style. However, more than this metric is required. Motivated by \citeauthor{jin2022deep}, we evaluate style transfer across the three main properties of text style transfer: style transfer accuracy, content preservation, and fluency. We use our custom joint sequence classification models, implemented using HuggingFace libraries \cite{wolf2020transformers} to evaluate the style transfer success ratio. Our definition for the Style Transfer Success $S_c$ is the total number of matches between intended and transferred style buckets, divided by the total number of samples. To judge content preserved in style transferred text, we use three metrics: BLEU \cite{papineni-etal-2002-bleu}, embedding-based similarity \cite{wieting2019beyond} using cosine similarity of two sentence embeddings \cite{sentencetransformers}, and Word Mover’s Distance (WMD) \cite{styletransfereval}. For fluency, we use measures like perplexity using GPT2 \cite{radford2019language} and an adversarial classifier using the cross-aligned autoencoder model \cite{styletransfereval}.

\noindent \textbf{Experimental Setup:} In this paper, we illustrate different micro-style combinations in the training data, for a randomly selected case, with each combination in both the ``balanced`` and ``skewed `` settings. Therefore, we consider 6 cases respectively: 1) Formality and Arousal in a balanced setting (\texttt{FA balanced}) 2) Formality and Arousal in a skewed setting (\texttt{FA skewed}) 3) Formality, Arousal and Bias in a balanced setting (\texttt{FAB balanced}) 4) Formality, Arousal and Bias in skewed setting (\texttt{FAB skewed}) 5) Formality, Arousal, Bias and Sentiment in the balanced setting (\texttt{FABS balanced}) 6) Formality, Arousal, Bias and Sentiment in skewed setting (\texttt{FABS skewed}). We construct the training data with the appropriate settings and then pass them through our experimental pipeline (illustrated in Figure~\ref{fig:fig2}) and quantitatively evaluate the style transfer results.

\noindent \textbf{Discussion:} Table \ref{tab:qual} shows examples of style-transferred sentences, given a style-transfer goal from our experimental pipeline for both balanced and skewed settings. E.g., given the objective is to decrease Formality but increase arousal, the sentence `` Did you hear about the soldier with 8 limbs? He was army'' transforms to ``He's an army soldier with 8 legs?''. Here, the contraction ``He's'' indicates a formality decrease, and the replacement of limbs with legs indicates a decrease. The overall arousal of this sentence is higher when it transforms into a question.

\begin{figure}[t]
\centering
\includegraphics[width=0.45\textwidth]{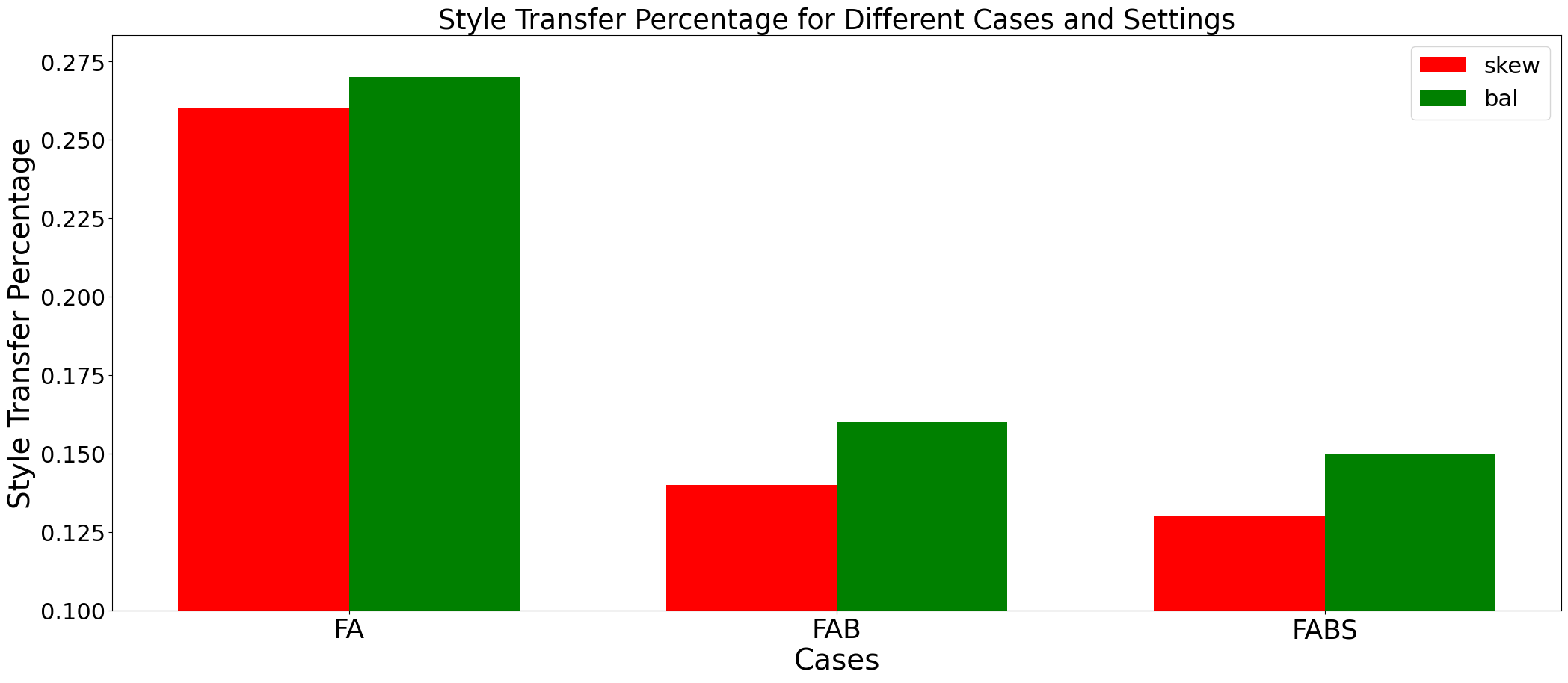} 
\caption{Balancing micro-style distributions leads to a higher multi-style transfer percentage than in the Skewed setting in all the cases.}
\label{fig:bal}
\end{figure}

Figure \ref{fig:bal} illustrates that the \textit{balanced setup always has a higher success percentage of style transfer ($S_c$) than the skewed setup}. We cannot compare the success percentage across cases because matching the exact target and transferred style buckets becomes difficult as the number of micro-styles increases. We can also observe through Table \ref{tab:qual} that the \textit{quality of the balanced transferred text aligns better with the style transfer goal than the skewed transferred text}.

\begin{figure}[t]
\centering
\includegraphics[width=0.5\textwidth]{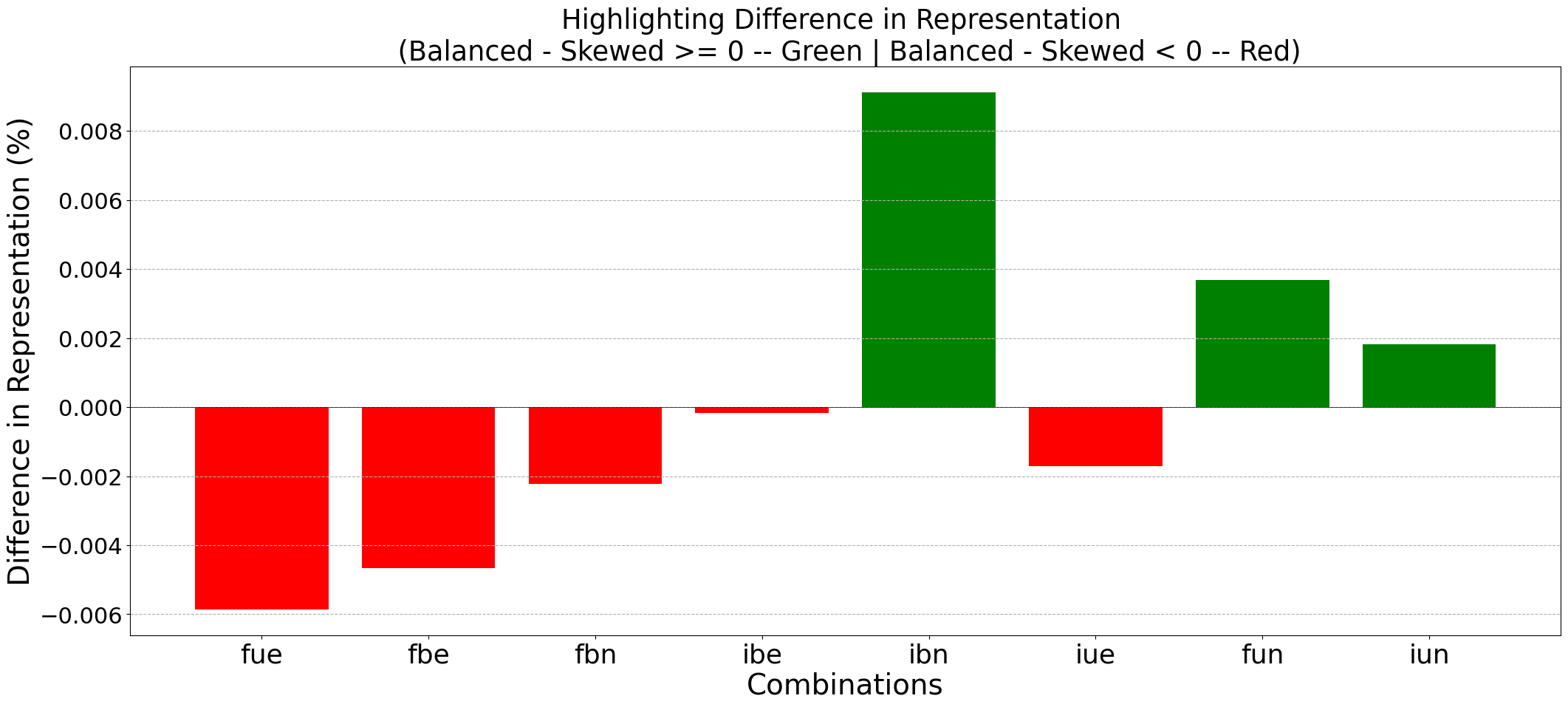} 
\caption{Considering the micro-style combinations such that, Formality [formal = f, informal = i], Bias [biased = b, unbiased = u], and Arousal [aroused = e, un-aroused = n], we observe that the micro-style combinations that are rarer (e.g., informal unbiased neutral (iun)) have more representation in the  ``balanced'' setting than the ``skewed'' setting. }
\label{fig:fig4}
\end{figure}

In Figure~\ref{fig:fig4}, we compare the difference in representation percentage of specific style combinations in the test sample for a specific case where we consider Formality, Arousal, and Bias micro-styles. We observe that a \textit{balanced joint distribution leads to more representation in the style combinations that are less likely to occur}. This is further accentuated as micro-styles increase, as reported in Appendix \ref{sec:app2}. In Figure~\ref{fig:fig4}, we see that rarer style combinations [\texttt{ibn}, \texttt{fun}, \texttt{iun}] show more representation in the balanced case as compared to the skewed case. This supports our intuition that the style transfer model benefits from learning the representation of all possible style combinations that can occur together.

When we consider Formality, Arousal, and Bias micro styles together, the most represented category (30\% of samples) is ``formal unbiased aroused'' (\texttt{fue}). The least represented category (as unlikely to occur together) is ``informal unbiased unaroused'' (\texttt{iun}) with 1\%. We observe that the quantitative evaluation metrics are quite indicative when compared across style combinations. For instance, in Table ~\ref{tab:tab2}, we observe that \textit{perplexity increases in categories that are unlikely to occur together} (\texttt{iun}). This indicates that the style transfer model is confused by the style distributions present for this style combination. 

We do not claim that our method of balancing multiple styles will work even for entangled micro-style combinations, as that is out of the scope of the current paper. However, balancing considerably affects the multi-style transfer output for the range of micro-style combinations we considered, and that has an application in many NLP tasks. This result could hugely influence future studies exploring better ways to balance even the entangled micro-styles.

\begin{table}[t]
\scriptsize
\centering
\begin{tabular}{@{}M{1.3cm}|M{0.8cm} | M{0.8cm} M{0.4cm} M{0.7cm} M{0.3cm} M{0.6cm}@{}}
 \toprule
 \textbf{Setting} & \textbf{Styles} & \textbf{Perp} & \textbf{Adv} & \textbf{BLEU} & \textbf{Cos} & \textbf{WMD} \\
  \midrule
  \multirow{2}{*}{Balanced} & \texttt{fue} & \textbf{115.16} & 0.90 & 0.77 & 0.92 & 0.32 \\


  & \texttt{iun}* & \textbf{598.58} & 0.86 & 0.78 & 0.92 & 0.36 \\
  \midrule
  \multirow{2}{*}{Skewed} & \texttt{fue} & 116.02 & 0.90 & 0.78 & 0.92 & 0.32 \\
  
  
  & \texttt{iun}* & 650.47 & 0.82 & 0.77 & 0.93 & 0.37 \\
 \bottomrule
\end{tabular}
 \caption{\label{tab:tab2} Comparison of the evaluation metrics for the most represented style combination (\texttt{fue} - formal unbiased aroused) vs the least represented style combination (\texttt{iun}* - informal unbiased unaroused). One key observation is that perplexity increases when the style combinations are unlikely to occur together.  }
\end{table}

\section{Conclusion}

Multi-style text style transfer is a challenging problem predominantly plagued by the need for jointly annotated high-quality datasets. There is a clear need for more research about the marginal and joint distribution of inherent micro-styles present in the training dataset used for style transfer. Multi-style text-style transfer typically requires access to large, jointly labeled datasets and many computational resources under typical implementations. More importantly, we would not be able to conveniently tweak the input data distributions in other multi-style text style transfer methods. 

In this paper, we implement a multi-style transfer pipeline that subverts the requirement of a jointly annotated dataset of multiple styles by constructing a pseudo-parallel dataset to which we introduce our contribution of constructing style distributions. We then use the modified pseudo-parallel datasets for multi-style transfer. Our modified pipeline effectively allows us to understand the importance of the joint distribution of micro styles in training data and is a substantial contribution. 

We quantitatively explore the impact of joint micro-style distributions in the training dataset on the style-transferred output sentences. When the joint micro-style distributions are balanced, there is more control over style-transferred output than with a skewed distribution. These findings will likely inform the design of multi-style transfer datasets and encourage us to explore the micro-style relationships in our datasets.

\section*{Limitations} 
In this research, though we employed automatic evaluation of our multi-style transferred text, we acknowledge that multi-style transfer is challenging to observe with the existing metrics for style transfer evaluation, and human evaluation should be done as well. As this research paper focuses on exploring the impact of style distributions in the training data on style-transferred output rather than developing a superior multi-style text transfer model,  we use quantitative evaluation in this iteration of our paper. We hope that the large sample size and the consistency of applied metrics make our automated approach a reasonable way of evaluating the style transfer output. 

This iteration of our paper aims to achieve multi-style transfer across multiple micro styles taken into consideration together as our contribution would aid in constructing a training dataset for multiple micro-style style transfers. We did not explore another exciting question of how balancing multiple micro styles in the training dataset might influence individual style transfer, which could be a promising future direction for our study.

We acknowledge that the classifier's quality sets an upper bound on the best style transfer accuracy that is obtainable. However, the target task is quite complicated without a parallel dataset. Our objective was not to have the most accurate classification of micro styles but to find a means to get acceptable pseudo labels for the micro styles. Individually, all our micro style classifiers had a classification accuracy of 80\% F1 and higher, and we deemed this good enough for pseudo-label creation. 

We also focused on utilizing the present styles in the training data and classifying them to derive inherent training style distributions instead of dynamically tuning the proportion of styles present in the training dataset. However, tuning these style proportions using techniques such as PPLM \cite{dathathri2019plug} would give us greater control over our experimental pipeline and is an appropriate next step.
 
\section*{Acknowledgement}
We thank Vivek Aithal, Priyam Srivastava and Daniel McAndrew for their initial work on the pipeline for multi-style transfer. This was instrumental to our project and helped us get a kickstart on our research.

\bibliography{custom}
\bibliographystyle{acl_natbib}

\appendix
\section{Dataset Information}
\label{sec:dataset}
We choose four micro-styles from both intended and unintended style categories, based on the style hierarchy as defined in \citeauthor{troiano2021theories} - Formality, Arousal, Sentiment, and Bias.
While formality is considered a ``non-targeted intended'' micro-style, arousal and sentiment are ``targeted intended'' micro-styles. We also include subjective bias, an ``unintended'' micro-style, to ensure we include styles from all hierarchy branches. We built our micro-style joint classification and style transfer models from multiple publicly available NLP datasets built by other researchers, and we detail these below.

\textbf{Formality}.
We use Grammarly’s Yahoo Answers Formality Corpus \cite{rao2018dear}, which consists of 105k sentences from two styles: ``formal'' and ``informal'' sentences written either in formal or informal modern English. Unlike formal sentences, informal sentences tend to have more misspellings, short forms (``u'' instead of ``you''), and non-standard usage of punctuation. 

\textbf{Arousal}.
We use the emotion annotated Emobank dataset \cite{buechel2022emobank} based on the three-dimensional VAD model developed by \cite{warriner2013norms}. In particular, we transform the Arousal dimension into binary categories such as ``arousal" and ``non-arousal." 

\textbf{Sentiment}. 
We use the famous Sentiment140 dataset \cite{go2009twitter}, which consists of automatically annotated tweets, where the tweets containing positive emoticons are assumed as positive. In contrast, those with negative emoticons are assumed to be negative. The training dataset consisted of 1.6M tweets, and the test dataset consisted of 359 tweets. The tweets were preprocessed using NLTK to remove special Twitter-specific symbols like hashtags, usernames, and URLs. 

\textbf{Bias}.
We use the Wiki Neutrality Corpus by \cite{pryzant2020automatically}. This is a new parallel corpus of 180,000 biased and neutralized sentence pairs.In order to train our joint classifier models, we used the training dataset from the appropriate micro-style datasets mentioned above. To implement our style distribution hypothesis, we used random samples for training and testing, from the combination of all the dev datasets from the benchmarks corpus by \cite{kang2019xslue}. This consists of 15 different styles coupled to both content and domain by varying degrees. We wanted to ensure that the dataset used for training our style transfer model and verifying our hypothesis has sufficient indicators of the appropriate micro-styles. This could be done best by using a sample consisting of datasets curated for each individual micro-style (since a jointly annotated dataset with so many styles is not available).

\section {Multi Style Transfer Pipeline}
\label{sec:app1}
\subsection{Resources used for Training}
All models were trained using cloud GPUs on Google Colab Pro and Pro+. We used  1 V100 GPU in its “High-RAM” (52GB) GPU run-time setting to train the paraphrase generation model, while for other models we used 1 P100 GPU at the ``standard RAM'' setting (32GB).
\subsection{Diverse Paraphrase Generation}
We adopted a pre-trained T5 model \cite{raffel2020exploring}, to generate paraphrases. We trained the model on the ParaNMT-filtered dataset provided by \cite{krishna2020reformulating}. This is a subset of the ParaNMT dataset with filters applied to promote lexical diversity, syntactic diversity, and semantic similarity. This model was then used to generate the pseudo-parallel training data for transfer. We selected the t5-small architecture (60 million parameters) as this is approximately 10x smaller than the GPT-2 large model used in \cite{krishna2020reformulating}. We used the hyper-parameters given in Table \ref{tab:apptable0}. Based on the recommendation in the appendix of Raffel et al, we used the “paraphrase: ” prefix to train the paraphraser model. Once we had this trained paraphrase model, we used diverse beam search \cite{vijayakumar2016diverse} to generate diverse paraphrased outputs. The hyper-parameters used for diverse beam search are mentioned in Table \ref{tab:apptable1}. We preferred beam search over top-p sampling in order to prioritize fluent paraphrases \cite{welleck2019neural} over unique paraphrases. \\

\noindent
\texttt{Input} - paraphrase: I love to play my guitar and I do not know why \\
\texttt{Output} - I love playing my guitar and I’m not sure why

\begin{table}[H]
    \centering
    \begin{tabular}{ |p{3cm}|p{3cm}| }
         \hline
         Hyperparameters & Value \\
         \hline
         batch size & 8 \\
         number of epochs   & 12  \\
         learning rate &   1e-4  \\
         max sequence length & 64 \\
         \hline
    \end{tabular}
    \caption{Hyper parameters for T5 training for paraphrase generation}
    \label{tab:apptable0}
\end{table}

\begin{table}[H]
    \centering
    \begin{tabular}{ |p{3cm}|p{3cm}| }
         \hline
         Hyperparameters & Value \\
         \hline
         max length & 70 \\
         early stopping   & True  \\
         no repeat ngram size &   5  \\
         num beams & 9 \\
         num beam groups & 3 \\
         diversity penality & 0.5 \\
         \hline
    \end{tabular}
    \caption{Hyperparameters for Beam Search}
    \label{tab:apptable1}
\end{table}

\subsection {Micro-style Classification}
We trained a joint sentence classification model to classify the sentence on multiple axes inspired by the approach in \citeauthor{kang2019xslue}, which uses an encoder-decoder-based model that learns cross-style patterns with the shared internal representation across styles. Our joint model comprises fully connected layers attached to a DistilBERT model \cite{sanh2019distilbert}, which acts as an encoder. The hyperparameters for this joint model are given in Table \ref{tab:apptable2}. This single model effectively replaces the need for a different model for each classification task, significantly reducing the need for computing resources for training and inference. Our joint classifier is essential for downstream tasks like training style transfer models and evaluation. Say we first perform a joint classification of both formality and arousal micro-styles on our datasets, considering we want a multiple-style transfer along the axes of formality and arousal. This results in both formality and arousal pseudo-labels for the sentences. Since these labels are generated algorithmically rather than by hand, we refer to them as \textit{pseudo-labels}. Pseudo-labeled sentences can then be used to generate the pseudo-parallel dataset for training joint style transfer models and directly measure the variation of a style along the axis of interest.\\

\begin{table}[H]
    \centering
    \begin{tabular}{ |p{3cm}|p{3cm}| }
         \hline
         Hyperparameters & Value \\
         \hline
         train batch size & 256 \\
         test batch size & 512 \\
         number of epochs   & 3  \\
         learning rate &   1e-4  \\
        
         \hline
    \end{tabular}
    \caption{Hyperparameters for Joint Classifier}
    \label{tab:apptable2}
\end{table}

\subsection{Pseudo Parallel Data Generation}
We then selected the best paraphrase (most stylistically different from the anchor sentence) based on the cosine distance between the anchor and the paraphrased sentence's style vectors. To enable the transfer model to transfer to specified levels of a particular style, we defined 'very low', 'low', 'mid', 'high', and 'very high' buckets for each micro-style. In the following, we describe the bucket boundaries for our style scores.

\noindent
\texttt{Buckets}: 
Very Low = [0, 0.2] Low = [0.2, 0.4] 
Mid = [0.4, 0.6]  
High = [0.6, 0.95] Very High = [0.95,1]

Using the absolute difference between original text style scores and their best-paraphrased text style scores, we find paraphrasing successfully stripped away both formality and arousal aspects of the text. The same phenomenon has been observed in previous studies, such as \cite{krishna2020reformulating}. 
To ensure a diverse pseudo-parallel dataset, we retain only anchor-paraphrase pairs that do not match in terms of their style bucket. For example, if an anchor-paraphrase sentence pair is assigned style buckets for formality and arousal, as [very high, low] and [very high, very low], this pair will be retained. However, if both style buckets match, the sentence pair will not be considered diverse enough to remain in the pseudo-parallel dataset.  In style transfer models, the paraphrased sentence and its style buckets are used as input parameters, while the style buckets assigned to the anchor sentence are used as proxy levels of output style transfer. The following is an example of the input to the joint style transfer model and the expected output. \\

\noindent
\texttt{Goal} - Highly increase the formality of the sentence, slightly increase the arousal of the sentence \\
\texttt{Input} - transfer: I’m sad you’re going $\vert$ input formality: low $\vert$ input arousal: low $\vert$ output formality: high $\vert$ output arousal: mid \\
\texttt{Output} - I am sorry you are going to go.

\begin{table}[H]
    \centering
    \begin{tabular}{ |p{3cm}|p{3cm}| }
         \hline
         Hyperparameters & Value \\
         \hline
         train batch size & 8 \\
         test batch size & 8 \\
         number of epochs   & 5  \\
         learning rate &   1e-4  \\
        
         \hline
    \end{tabular}
    \caption{Hyperparameters for T5 for Style Transfer}
    \label{tab:apptable3}
\end{table}
\begin{figure*}
\centering
\includegraphics[width=0.8\textwidth]{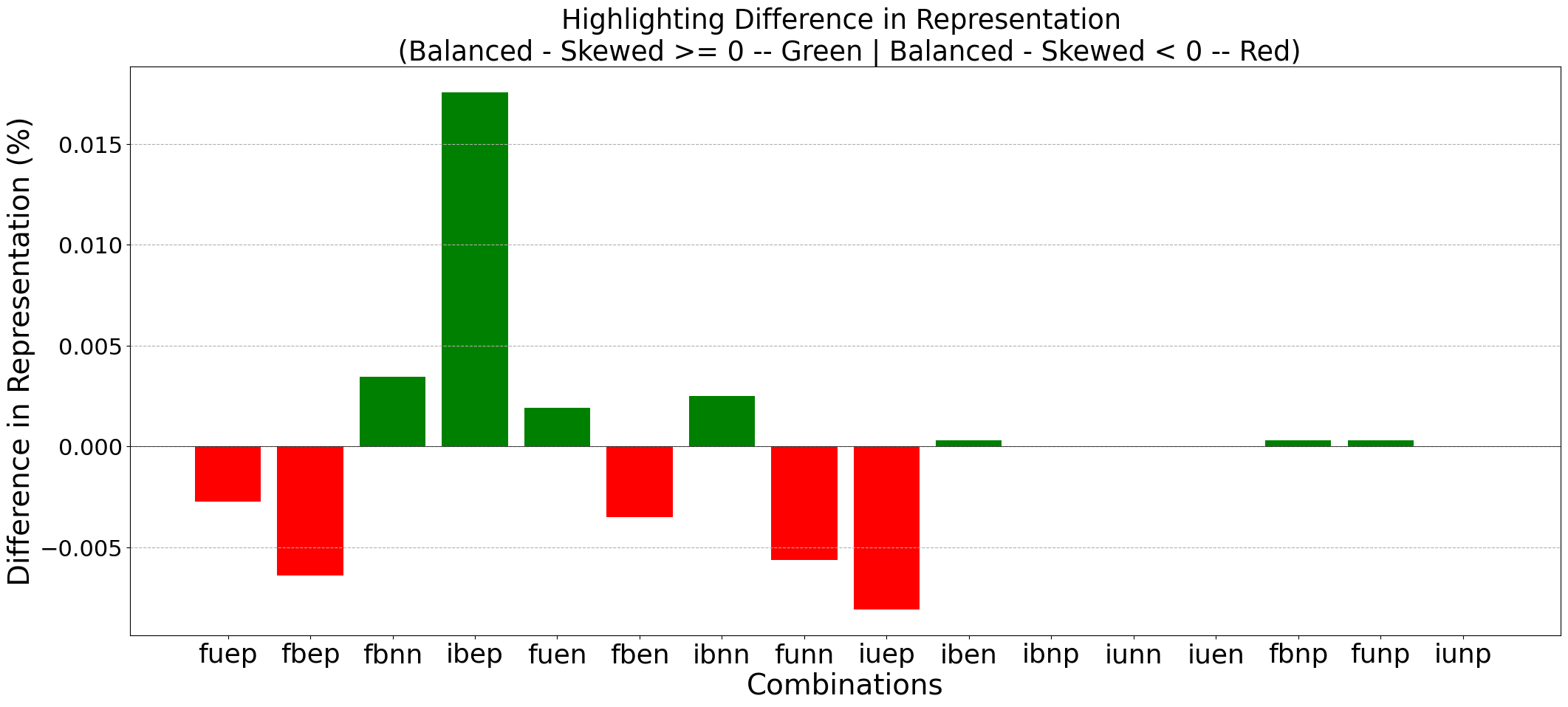} 
\caption{Considering the micro-style combinations such that, Formality [formal = f, informal = i], Bias [biased = b, unbiased = u], Arousal [aroused = e, un-aroused = u], and Sentiment [negative = n, positive = p]; we observe that the micro-style combinations that are rarer have more representation in the  ``balanced'' setting than the ``skewed'' setting. The categories \texttt{fbnp}, \texttt{funp} and \texttt{iben} have more representation for balanced setting vs skewed setting. }
\label{fig:appfig2}
\end{figure*}
\subsection {Style Transfer Training}
Our T5 models were trained on pseudo-parallel datasets created and filtered as described earlier. According to the task, we converted the classifier predictions into buckets of style based on the chosen style of the original and then paraphrased sentences. To achieve the necessary intensity transfers, we appended this information to the paraphrased sentence, as motivated by the original T5 paper \cite{raffel2020exploring}. Hyperparameters are mentioned in Table \ref{tab:apptable3}. For generating sentences from our trained models, we used a combination of both beam search \cite{vijayakumar2016diverse} and nucleus sampling \cite{holtzman2019curious} and chose the top 3 sentences from the generations.

\section{Some Additional Results}

\label{sec:app2}
\subsection{Impact of Fluency filter on input training data}

We find that filtering the original dataset based on fluency metrics always results in better style transferred output as compared to the transferred output when the input dataset is not filtered. This is intuitive, as better quality input prevents confusion in the style transfer model and leads to better quality output. As a result of this finding, we use a fluency filter (adversarial classifier $> 0.1$ and perplexity $< 365$), before we conduct any of the rest of our experiments with micro-style distributions.

\subsection{Balancing effect on lesser represented style combinations}

\begin{figure}[t]
\centering
\includegraphics[width=0.5\textwidth]{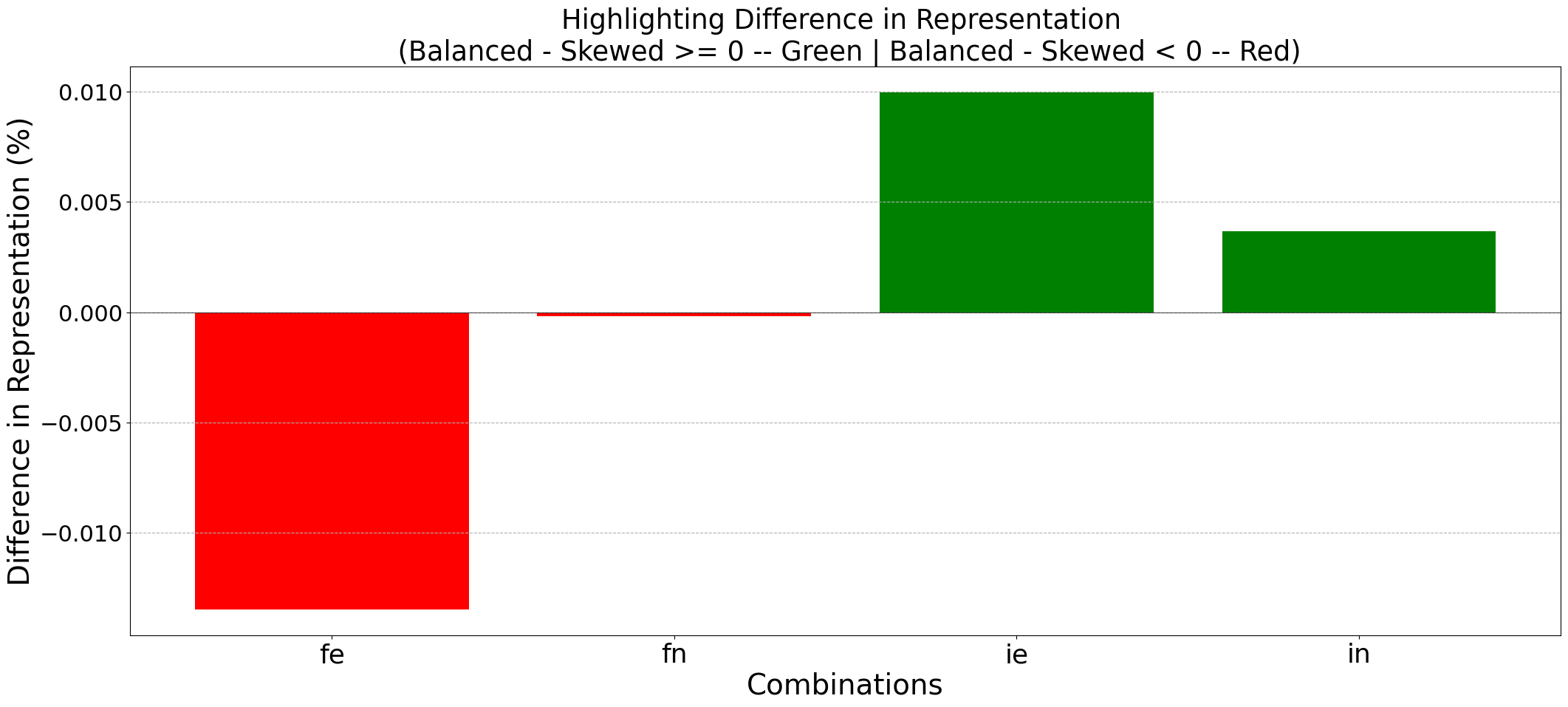} 
\caption{Considering the micro-style combinations such that, Formality [formal = f, informal = i]and Arousal [aroused = e, un-aroused = u]; we observe that the micro-style combinations that are rare (\texttt{ie}, \texttt{in}) have more representation in the  ``balanced'' setting than the ``skewed'' setting. }
\label{fig:appfig1}
\end{figure}

In Figure \ref{fig:appfig1},  we consider the case where we examine Formality [formal = f, informal = i] and Arousal [aroused = e, un-aroused= u] micro-styles and compare the percentage of specific style combinations in the test sample. We observe that as the number of micro styles increases, a balanced joint distribution leads to more representation in combinations that are less likely to
occur such as \texttt{in} or 'informal and neutral'. 

Figure \ref{fig:appfig2} shows a similarly pronounced effect. Here the number of micro styles is increased, and we can observe that the balanced setting shows higher representation than the skewed setting. An example of an unlikely style combination is \texttt{fbnp}, or ``formal biased neutral and positive''. We also observe that as the number of micro-styles increases, there is no representation in some combinations in both settings [\texttt{ibnp,iunn,iuen,iunp}]. This is a natural result as some micro-style combinations cannot exist in nature.\\

\end{document}